\documentclass[12pt,a4paper]{article}
\usepackage{amsfonts}  
\usepackage{amsthm} 
\usepackage{amsmath}  
\usepackage{amssymb} 
\usepackage{mathrsfs}
\usepackage{dsfont}   
\usepackage{latexsym,epsf,graphicx}
\usepackage{caption}
\usepackage[colorlinks=false]{hyperref}
  \hypersetup{ 
    pdftitle={A note on extension theorems and its connection to universal consistency in machine learning},
    pdfauthor={\textcopyright Andreas Christmann}
   }
\usepackage[dvips]{color}
\usepackage{url}
\usepackage[sort&compress]{natbib}  
 \bibpunct{(}{)}{;}{a}{,}{,}
\newcommand{\bc}{\begin{center}}
\newcommand{\ec}{\end{center}}
\newcommand{\bi}{\begin{itemize}}
\newcommand{\ei}{\end{itemize}}
\newcommand{\be}{\begin{equation}}
\newcommand{\ee}{\end{equation}}
\newcommand{\beqna}{\begin{eqnarray*}}
\newcommand{\eeqna}{\end{eqnarray*}}
\newcommand{\bd}{\begin{displaymath}}
\newcommand{\ed}{\end{displaymath}}
\newcommand{\bt}{\begin{tabular}}
\newcommand{\et}{\end{tabular}}

\newcommand{\myem}[1]{\textbf{#1}}

\newtheorem{theorem}{Theorem}[section]

\newtheorem{definition}[theorem]{Definition}
\newtheorem{lemma}[theorem]{Lemma}
\newtheorem{example}[theorem]{Example}

\newtheorem{proposition}[theorem]{Proposition}

\newcommand{\pen}{\mathrm{pen}}

\newcommand{\rem}[1]{}
\newlength{\fixboxwidth}
\newcommand{\C}{\mathds{C}}

\newcommand{\N}{\mathds{N}}

\newcommand{\R}{\mathds{R}}

\newcommand{\K}{\mathds{K}}

\newcommand{\snorm}[1] {\Vert #1 \Vert}

\newcommand{\hnorm}[1] {\Vert #1 \Vert_{\sH}}
\newcommand{\hhnorm}[1] {\Vert #1 \Vert_{\sH}^2}
\newcommand{\inorm}[1] {\Vert #1 \Vert_\infty}

\DeclareMathOperator{\id}{id}

\newcommand{\Lx}[2] {{L}_{#1}(#2)}
\newcommand{\sH}    {H}  
\newcommand{\sA} {\mathcal{A}}     
\newcommand{\B}    {\mathcal{B}}   

\def \lb        { \lambda }

\def \g         { \gamma }
\def \e         { \varepsilon }

\def \s         { \sigma }

\def \d         { \delta }

\newcommand{\eins}{\boldsymbol{1}}

\newcommand{\kg}   {k_{\g}}

\def \P           { \mathrm{P} }   

\newcommand{\Ex}{\mathbb{E}}       

\def\zeitende{\hfill \quad \hbox{$\vartriangleleft$}}
\def\exampleende{\ifmmode\zeitende\else{\unskip\nobreak\hfil
\penalty50\hskip1em\null\nobreak\hfil\zeitende
\parfillskip=0pt\finalhyphendemerits=0\endgraf}\fi}

\newcommand{\cH}{\mathcal{H}}

\newcommand{\cF} {\mathcal{F}}
\newcommand{\cR}{\mathcal{R}}
\newcommand{\cX}{\mathcal{X}}
\newcommand{\cY}{\mathcal{Y}}

\newcommand{\cW}{\mathcal{W}}

\newcommand{\cXY}{{\cX\times\cY}}

\newenvironment{declaration}[1]{\trivlist \item[\hskip \labelsep{\em #1 }]\ignorespaces}{\endtrivlist}
\newenvironment{proofof}[1]{\begin{declaration}{#1.}}{\end{declaration}}

\def \O { \Omega }

\newcommand{\bnum}{\begin{enumerate}}
\newcommand{\enum}{\end{enumerate}}

\def \lb        { \lambda }

\newcommand{\beq}{\begin{eqnarray}}
\newcommand{\eeq}{\end{eqnarray}}

\newcommand{\qedr}{\hfill \quad \qed}

\def\textregtrademark{\raise1ex\hbox{\scriptsize\textregistered}} 

\numberwithin{equation}{section}

\title{\textbf{A short note on extension theorems and their connection to universal consistency in machine learning}$^\dag$\footnotetext{\dag
Corresponding author: A. Christmann, Email: \url{andreas.christmann@uni-bayreuth.de}\newline
~The work by A. Christmann described in this paper is partially supported by a grant of the Deutsche
Forschungsgesellschaft [Project No. CH/291/2-1]. The work by 
D.~H. Xiang is supported by the National Natural Science Foundation of
China under Grant 11471292 and the Alexander von Humboldt Foundation of
Germany.}}
 
\author{\textbf{Andreas Christmann$^1$, Florian Dumpert$^1$, Dao-Hong Xiang$^{2,1}$}\\
\normalsize{$^1$ Department of Mathematics, University of Bayreuth, Germany}\\
\normalsize{$^2$ Department of Mathematics, Zhejiang Normal University, 
Jinhua, Zhejiang 321004, China} }
\date{Date: \today}

\begin{document} 

\maketitle
 
\begin{abstract}
Statistical machine learning plays an important role in modern statistics
and computer science. One main goal of statistical machine learning is to
provide universally consistent algorithms, i.e., the estimator converges in
probability or in some stronger sense to the Bayes risk or to the Bayes decision function.
Kernel methods based on minimizing the regularized risk over a reproducing kernel
Hilbert space (RKHS)  belong to these statistical machine learning methods.
It is in general unknown which kernel yields optimal results for a particular
data set or for the unknown probability measure. Hence various kernel learning methods were proposed to choose
the kernel and therefore also its RKHS in a data adaptive manner.
Nevertheless, many practitioners often use the classical Gaussian
RBF kernel or certain Sobolev kernels with good success. 
The goal of this short note is to offer one possible theoretical explanation for this
empirical fact.
\end{abstract}  
  
\noindent{\bf Key words and phrases.} Machine learning; kernel learning; universal consistency; Dugundi extension theorem; Lusin theorem; dense; reproducing kernel Hilbert space.



\section{Introduction}\label{intro} 

Regularized empirical risk minimization over large classes $\cF$ of 
functions $f: \cX\to \cY$ have attracted a lot of interest during the last decades in statistical machine learning. Here $\cX$ and $\cY$ denote the so-called input space and output space, respectively.
Of particular importance is the case that $\cF$ equals a reproducing kernel Hilbert space $H$ specified by its corresponding kernel $k$.
Fields of applications range from classification,  
regression, and quantile regression to ranking, similarity learning and
minimum entropy learning. 
Probably the most important goal of such machine learning methods is
universal consistency, i.e., convergence in probability to the Bayes risk 
defined as the infimum of the risk over all measurable functions or to the
Bayes decision function, if it exists.
To achieve this goal, one typically splits up the total error
into a stochastic error and into a non-stochastic approximation error.
Concentration inequalities are then often used to upper bound the stochastic error.
Denseness arguments are typically used to show that the
infimum of the risk when minimizing over $\cF$ equals the Bayes risk, i.e.,
\be 
   \inf_{f\in\cF} \cR_{L,\P}(f) = \inf_{f\,\mathrm{measurable}} \cR_{L,\P}(f),
\ee
where $L$ denotes a loss function and $\P$ denotes a probability measure.
Two of the most successful special cases in statistical machine learning are the following ones.
Let the input space $\cX$ be a compact metric space. Then a continuous kernel is called
universal, if its RKHS $H$ is dense with respect to the supremum norm in $C(\cX)$, see Definition \ref{Def.universalkernel}. For more general input spaces, one often assumes
that the RKHS is dense in some $L_p(\mu)$ for all probability measures $\mu$ on 
$\cX$,
where $p\ge 1$ is some constant, see e.g., \citet[Lem.\,4.59, Thm.\,4.63]{SC2008}.
 
The main goal of this paper is to address the question whether denseness of the 
RKHS with respect to the supremum norm in $C(\cX)$ can sometimes be weakened.
 
To fix ideas, let $n$ be a positive integer and $D=((x_1,y_1), \ldots, (x_n,y_n))$ be a given data set,
where the value $x_i\in \cX$ denotes the input value and $y_i\in\cY$ denotes the output value
of the $i$-th data point. 
Let 
$$
  L:\cX\times\cY\times\R\to[0,\infty)
$$
 be a loss function of the form  
$L(x,y,f(x))$, where $f(x)$ denotes the predicted value for $y$, if $x$ is observed, and $f:\cX\to\R$ is a 
real-valued function.
Most often $L$ is assumed to be a convex loss function, i.e., $L(x,y,\cdot)$
is convex for any fixed pair $(x,y)\in\cXY$.
Many regularized learning methods are then defined as minimizers of the optimization problem
\be \label{RERM}
\inf_{f\in\mathcal{F}} \frac{1}{n} \sum_{i=1}^n L(x_i,y_i,f(x_i)) + \pen(\lb_n,f),
\ee
where the set $\mathcal{F}$ consists of functions $f:\cX\to\R$, $\lb_n>0$ is a regularization
constant, and $\pen(\lb_n,f)\ge 0$ is some regularization term to avoid overfitting for the case, that
$\mathcal{F}$ is rich. One example is that $\mathcal{F}$ is a reproducing kernel Hilbert space $H$
and $\pen(\lb_n,f)=\lb_n\hhnorm{f}$, i.e., 
\be \label{SVMs}
\inf_{f\in{H}} \frac{1}{n} \sum_{i=1}^n L(x_i,y_i,f(x_i)) + \lb_n \hhnorm{f},
\ee
see e.g., 
\cite{Vapnik1995,Vapnik1998}, \cite{PoggioGirosi1998}, \cite{Wahba1999}, 
\cite{ScSm2002}, \cite{CuckerZhou2007}, 
\cite{SmaleZhou2007}, \cite{SC2008} and the references cited therein. 
If the output space $\cY$  is a general Hilbert space, regularized learning with
kernels have been investigated, e.g., by 
\citet{MicchelliPontil2005b} and 
\citet{CaponnettoDevito2007}. 
We also like to mention two other regularization terms:  
$\pen(\lb_n,f):=\lb_n \hnorm{f}^p$ for some $p\ge 1$ and  elastic nets, see e.g., \citet{DemolDevitoRosasco2009}. 

In recent years there is increasing interest in related pairwise learning methods where a 
\emph{pairwise loss function} 
$$
  L:\cX\times\cY\times\cX\times\cY\times\R\times\R\to[0,\infty)
$$ 
is used and  optimization problems like
\be \label{PairwiseLearning}
\inf_{f\in H} \frac{1}{n^2} \sum_{i=1}^n \sum_{j=1}^n L(x_i,y_i,x_j,y_j,f(x_i),f(x_j)) + 
\lb_n \hhnorm{f}
\ee
have to be solved.
An example of this class of learning methods occurs when one is interested in minimizing Renyi's entropy of order 2, see e.g., \cite{HuFanWunZhou2013}, 
\cite{FanHuWuZhou2014}, and \cite{YingZhou2015} for consistency and fast learning rates.
Another example arises from ranking algorithms, see e.g., \cite{Clemencon2008} and \cite{AgarwalNiyogi2009}. Other examples include gradient learning, and metric and similarity learning, see e.g., \cite{MukherjeeZhou2006}, \cite{XinNgJordanRussell2002}, and \cite{CaoGuoYing2015}. 
We refer to \citet{ChristmannZhou2015b} for robustness aspects of pairwise learning
algorithms. 

In practice, the loss function is usually determined by the concrete application.
However, it is not always clear how to choose a kernel and therefore its RKHS in a reasonable manner.
There exist so many papers on learning the kernel from the data --often called 
kernel learning or multiple kernel learning-- that it is impossible to cite
all them here, but we like to mention a few.
One popular approach is to consider a linear or a convex combination of several 
fixed kernels or of their corresponding reproducing kernel Hilbert spaces.
\citet{LanckrietEtAl2004} proposed to learn the kernel matrix with semidefinite
programming and
\citet{MicchelliPontil2005} proposed to learn the kernel function via regularization.
\citet[Thm. 3]{YingZhou2007} proposed learning with Gaussian RBF-kernels with flexible bandwidth
parameters.
A direct method for building sparse kernel learning algorithms was proposed by \citet{WuSchoelkopfBakir2006}.
Large scale multiple kernel learning was investigated by \citet{SonnenburgEtAl2006}.
\citet{Bach2008} considered consistency of the group lasso and multiple kernel learning.
We also refer to \citet{RakotomamonjyEtAl2008} and \citet{GoenenAlpaydin2011} for multiple kernel learning algorithms
and to \citet{KoltchinskiiYuan2010} for sparsity considerations of
such algorithms.
Learning rates of multiple kernel learning with $L_1$ and elastic-net regularizations and a trade-off between sparsity and smoothness were considered by \citet{SuzukiSugiyama2013}.
A different approach was given by \citet{OngSmolaWilliamson2005},
who proposed learning the kernel via hyperkernels. The idea behind hyperkernels
is to consider the kernel $k:\cX\times \cX\to\R$ as an unknown function 
from $\cX^2$ to $\R$. To estimate $k$ from the data set, a second
kernel $\tilde{k}:\cX^2\times \cX^2\to\R$ is constructed and 
optimization over the RKHS corresponding to $\tilde{k}$ is done.

The rest of the paper has the following structure.
To improve the readability of this paper, 
we list some well-known results on kernels and reproducing kernel Hilbert spaces in Section \ref{kernel} and on extension theorems and Lusin's theorem in Section \ref{extensionlusin}.
Section \ref{result} contains with Theorem \ref{thm.convergence} our result.

\section{Kernels}\label{kernel}

Kernels and reproducing kernel Hilbert spaces (RKHSs) play a central role in modern nonparametric statistics and
machine learning. We refer to
\citet{BergChristensenRessel1984},
\citet{BenyaminiLindenstrauss2000}, and
\citet{BerlinetThomasAgnan2004} and references therein for details.
Here we focus only on some aspects of RKHSs which are important for the present paper.
Let $\K\in\{\R,\C\}$ and $\cX$ be a non-empty set.
A function $k:\cX\times \cX\to \K$ is called a
\emph{kernel} on $\cX$
if there exists a $\K$-Hilbert space $H$ and a map $\Phi:\cX\to H$ such that
for all $x,x'\in \cX$ we have
\begin{equation}\label{kernel-def-eq}
k(x,x') \ =\  \langle\Phi(x'),\Phi(x)\rangle_H\, .
\end{equation}
We call $\Phi$ a \emph{feature map} and $H$ a \emph{feature space} of $k$.
A $\K$-Hilbert function space $H$ consists of functions mapping from $\cX$ into $\K$.

\begin{definition}\label{kernel-rkhs}
Let $\cX\ne \emptyset$ and  
$H$ be a $\K$-Hilbert function space over $\cX$.
\begin{enumerate}
\item A function $k:\cX\times \cX\to \K$ is called a \myem{reproducing kernel} of $H$
    if we have $k({\,\cdot\,},x)\in H$ for all $x\in \cX$ and the \myem{reproducing property}
    $$
    f(x) = \langle f, k({\,\cdot\,},x)\rangle_H
    $$
    holds for all $f\in H$ and all $x\in \cX$. The function 
    $\Phi: \cX\to H$, $\Phi(x):=k(\cdot,x)$
    is called \myem{canonical feature map} of $k$.
\item The space $H$ is called a \myem{reproducing kernel Hilbert space (RKHS)} over $\cX$
    if for all
    $x\in \cX$ the  Dirac functional $\d_x:H\to \K$ defined by 
    $\d_x(f):= f(x)$, $f\in H$, is continuous.
\end{enumerate}
\end{definition}
 
It is well-known that every Hilbert function space with a reproducing kernel is an
RKHS and that, conversely, every RKHS has a unique reproducing kernel which can be 
determined by the Dirac functionals.  
We will consider in the following only $\K=\R$,  because $\R$-valued kernels are most often used in practice. 

A kernel $k$ is \emph{bounded} if and only if
$\inorm{k}:= \sup_{x\in \cX}\sqrt{k(x,x)}<\infty$.

The Gaussian RBF kernel $k_\gamma$ defined on $\cX\subset \R^d$, where
$d\in\N$ and the width $\gamma>0$, is given by
\begin{equation}\label{Def.GRBF}
  k_\gamma(x,x')= \exp\left( -\frac{\|x-x'\|_2^2}{\gamma^2}\right), \qquad x,x'\in \cX.
\end{equation}
It is well-known that $k_\gamma$ is bounded and continuous and that hence all 
functions $f$ in its RKHS $H_\gamma$ are bounded and continuous, too.

The following notion of universal kernels was introduced by 
\citet[Def.\,4]{Steinwart2001a}. Please note the combination of a \emph{compact} metric
input space and a \emph{continuous} kernel.

\begin{definition}\label{Def.universalkernel}
A continuous kernel $k$ on a compact metric space $(\cX,d_\cX)$ is called 
\emph{universal} if the RKHS $H$ of $k$ is dense in $C(\cX)$ with respect to the
supremum norm,
i.e., for every function $f\in C(\cX)$ and every $\e>0$ there exists a function 
$g\in H$ with 
\be \label{universalkernelineq}
  \inorm{f-g} \le \e.
\ee
\end{definition}

Please note that the rather strong supremum norm is used in Definition
\ref{Def.universalkernel}.
This is to some extent surprising, because the universal consistency of 
learning algorithms is often defined by a weaker mode of convergence,
e.g., convergence in probability, 
to the Bayes risk or to the Bayes decision function. For kernel based regression, we refer e.g., to \citet{GyKoKrWa2002} for the least squares loss function and to \citet[Thm. 12]{ChristmannSteinwart2007a} for general convex  loss functions
of growth type $p\ge 1$. We refer to
\citet[Thm. 5, Thm. 6]{ChristmannSteinwart2008d} for kernel based quantile regression.

Universal kernels can separate compact and disjoint subsets of a compact metric space
as the following results shows. 

\begin{proposition}[{\citet[Prop.\,5]{Steinwart2001a}}]
Let $(\cX,d_\cX)$ be a compact metric space and $k$ be a universal kernel on $\cX$ with
RKHS $H$.
Then for all compact and mutually disjoint subsets $K_1,\ldots,K_n\subset \cX$,
all $\alpha_1,\ldots,\alpha_n\in\R$ and all $\e>0$ there exists a function $g$
induced by $k$, i.e., there exists $w\in H$ such that $g(x)=\langle w, k(\cdot,x) \rangle_H$ for all $x\in \cX$, with $\inorm{g}\le \max_i |\alpha_i|+\e$ such that
\begin{equation}
\left\| g_{|K} - \sum_{i=1}^n \alpha_i \eins_{K_i}\right\|_\infty \le \e \, ,
\end{equation}
where $K:=\bigcup_{i=1}^n K_i$ and $g_{|K}$ denotes the restriction of $g$ to $K$
and $\eins_{K_i}$ denotes the indicator function on $K_i$.
\end{proposition}
 
We refer to 
\citet{MicchelliXuZhan2006} 
and the references given therein for additional results on universal kernels
and relationships between their RKHSs and $C(\cX)$.
Special emphasis is given in that paper to 
\emph{translation invariant kernels} having the form
$k(x,x')=h(x-y)$ for continuous functions $h:\R^d\to\R$ 
and to \emph{radial
kernels} $k(x,x')=\phi(\|x-x'\|^2)$ on $\cX\subset\R^d$ for appropriate  functions 
$\phi:[0,\infty)\to\R$.
Such kernels were already investigated by 
\citet{Schoenberg1938}. We refer to
\citet{Wu1995} and \citet{Wendland1995} for radial kernels with compact support.
Many Wendland kernels have a Sobolev space as RKHS, for details we refer to 
\citet[Thm. 10.35]{Wendland2005}.

The next result on the denseness of RKHSs in some $L_p(\mu)$ spaces
is also useful to prove universal consistency results of kernel based methods, we refer e.g., to \citet[Thm. 4.26, Lem. 4.59]{SC2008}.

\begin{theorem}\label{kernel-lp-new}
Let $\cX$ be a measurable space, 
$\mu$ be a $\s$-finite measure on $\cX$, and $H$ be a separable RKHS over $\cX$ 
with measurable kernel $k:\cX\times \cX\to \R$.
Assume that there exists a $p\in [1,\infty)$ such that
$
\snorm k_{\Lx{p}\mu} := \bigl( \int_\cX k^{ p /2}(x,x) d\mu(x)\bigr)^{ 1 /p} < \infty\, .
$
Then
\begin{enumerate}
\item $H$ consists of $p$-integrable functions and the 
inclusion $\id:H\to L_p(\mu)$ is continuous with $\snorm{\id:H\to \Lx p \mu}\leq \snorm k_{\Lx p \mu}$.
\item The adjoint of this inclusion is the  operator $S_k:\Lx{p'}\mu\to H$ defined by 
\begin{equation}\label{kernel-lp-h2}
S_kg(x) := \int_\cX  k(x,x') g(x') d\mu(x')\, , \quad \qquad g\in \Lx{p'}\mu,\, x\in \cX,
\end{equation}
where $p'$ is  
defined by $\frac 1 p+\frac 1 {p'} = 1$. 
\item $H$ is dense in $L_p(\mu)$ if and only if $S_k:\Lx{p'}\mu\to H$  is injective.
\item If the operator $S_k$ defined in {(\ref{kernel-lp-h2})} is injective, then
      $H$  is dense in $\Lx q {h\mu}$ for all $q\in [1,p]$ and 
all measurable  $h:\cX\to [0,\infty)$ with $h\in \Lx s \mu$, where  $s:= \frac p{p-q}$.
\end{enumerate}
\end{theorem}

One can show that the operator $S_k$ is injective for any real-valued Gaussian RBF
kernel $k_\g$ given by {(\ref{Def.GRBF})}, which yields the following result, see e.g., 
\citet[Thm.\, 4.63]{SC2008}.

\begin{theorem}\label{kernel-rbf-dense}%
Let $\g>0$, $p\in [1,\infty)$, and $\mu$ be a finite measure on $\R^{d}$. Then the RKHS $H_{\g}(\R^{d})$
of the real-valued Gaussian RBF kernel $\kg$  is dense in $\Lx p \mu$. 
\end{theorem}
 
\citet[Cor. 4.9]{ScovelHushSteinwartTheiler2005} proved the following
more general result. Let $\cX\subset\R^d$, where
$\cX$ is not necessarily compact, and $k:\cX\times \cX\to\R$ be a non-constant 
radial kernel $k$. Then the RKHS of $k$ is dense in $L_p(\mu)$ for all
$p\in[1,\infty)$ and all finite measures $\mu$ on $\R^d$. Furthermore, 
if $\cX\subset\R^d$ is compact, then $k$ is universal.

\section{Extension theorems and Lusin's theorem}\label{extensionlusin}

To improve the readability of the paper, we now cite some facts from topology,
see e.g., \citet{Dugundji1966} or \citet{Dudley2002}.
A topological space $(\cX,\tau)$ is called \emph{normal} if for each pair of disjoint closed sets $E_1\subset \cX$ and $E_2\subset \cX$ there are disjoint open sets $O_i$ with $E_i\subset O_i$, $i\in\{1,2\}$. Every metric space and every compact
Hausdorff space are normal.
Recall, that a subspace of a normal space need not be normal. However, a closed subspace of a normal space is normal.

Let $(\cX,\tau_\cX)$ and $(\cW,\tau_\cW)$ be two topological spaces, $A\subset \cX$ closed, and $f: A \to \cW$ a continuous function. A continuous function $F: \cX \to \cW$ such that 
$F(a)=f(a)$ for all $a \in A$ is called an \emph{extension} of $f$ (over $\cX$ relative
to $\cW$). 
The classical Tietze (or Tietze-Urysohn) extension theorem shows that an extension of a \emph{real-valued} function $f$ is possible for normal spaces, see 
\citet[Thm. 2.6.4, p.\,65]{Dudley2002}. 

\begin{theorem}[Tietze-Urysohn extension theorem]\label{thm.TietzeUrysohn}
Let $(\cX,\tau_\cX)$ be a normal topological space and $A$ be a closed subset of $\cX$.
Then for any $c\ge 0$ and each of the following subsets $\cW$ of $\R$ with usual
topology, every continuous function $f: A\to \cW$ can be extended to a continuous
function $F: \cX\to \cW$: 
\bnum
\item $\cW=[-c, +c]$.
\item $\cW=(-c, +c)$.
\item $\cW=\R$.
\enum
\end{theorem} 

The following extension theorem was proven by \citet[Thm.4.1]{Dugundji1951}. 
This theorem makes a stronger assumption on $\cX$, but a weaker assumption on $\cW$.
Recall that a linear topological space is a vector space $\cW$ equipped with a
Hausdorff topology such that the two maps $\alpha: \cW\times \cW\to \cW$
and $m: \R\times \cW\to \cW$ (Euclidean topology on $\R$) are continuous,
see \citet[p.\,413]{Dugundji1966}. A linear topological space $\cW$ is \emph{locally convex} if for each $w\in \cW$ and 
neighborhood $\mathcal{U}(w)$ there is a convex neighborhood $\mathcal{V}$ such that $w \in \mathcal{V} \subset \mathcal{U}(w)$, see \citet[p.\,414]{Dugundji1966}.

\begin{theorem}[Dugundji extension theorem]\label{thm.Dugundji}
Let $(\cX,d_\cX)$ be a metric space, $A$ be a closed subset of $\cX$, 
$\cW$ be a locally convex linear topological space, and 
$f: A\to \cW$ a continuous map. Then there exists an 
extension $F: \cX \to \cW$ of $f$, i.e., $F: \cX\to \cW$ is a continuous function with $F(a)=f(a)$ for every $a\in A$. Furthermore, $F(\cX)$ is a subset of the convex hull
of $f(A)$.
\end{theorem} 

Of course, there is a close relationship between
Borel measurable functions and continuous functions:
if $f$ is a continuous map between metric spaces, then $f$ is Borel-measurable. 
However, it is well-known that there is a much deeper relationship between continuity and
Borel measurability. The following theorem is a generalisation of the classical Lusin theorem for real-valued functions to more general domain
and range spaces, see \citet[Thm. 7.5.2, p.\,244]{Dudley2002}.

\begin{theorem}[Lusin's theorem I]\label{thm.LusinI}
Let $(\cX,\tau)$ be any topological space and $\mu$ be a finite, closed regular
Borel measure on $\cX$. Let $(\cW,d_\cW)$ be a separable metric space and let 
$f: \cX\to \cW$ be a Borel measurable function. Then for any $\varepsilon>0$ there is a closed set $\cX_\varepsilon \subset \cX$ such that 
$\mu(\cX\setminus \cX_\varepsilon) < \varepsilon$ and the restriction of $f$ to 
$\cX_\varepsilon$ is continuous. 
\end{theorem} 

There exist other versions of Lusin's theorem for $\cX$ a Polish space or a locally compact space and compact sets  $\cX_\varepsilon \subset \cX$. Here, we only cite
the following result taken from 
\citet[Thm. 2.5.15, p. 187]{DenkowskiEtAl2003}.
Recall, that a topological space $(\cY,\tau_\cY)$ is called a Polish space, if the topology $\tau_\cY$ is metrizable by some metric $d_\cY$ such that $(\cY,d_\cY)$ is a complete separable metric space. 
 
\begin{theorem}[Lusin's theorem II]\label{thm.LusinII}
Let $(\cX,\tau)$ be a Polish space, $(\cW,d_\cW)$ be a separable metric space,
$f: \cX\to \cW$ be a Borel measurable function, and $\mu$ be a finite Borel measure on 
$(\cX,\B_\cX)$. Then for any $\varepsilon>0$ there 
is a compact set $\cX_\varepsilon \subset \cX$ such that 
$\mu(\cX\setminus \cX_\varepsilon) < \varepsilon$ and the restriction of $f$ to 
$\cX_\varepsilon$ is continuous. 
\end{theorem} 

One reason why Polish spaces are interesting in probability theory and statistics is the fact, than then regular conditional probabilities are uniquely defined,
see \citet[Thm. 10.2.2, p.\,345]{Dudley2002}. Furthermore, disintegration allows then to split a probability measure $\P$
defined on $(\cX\times \cY, \sA\otimes \B_\cY)$ into the marginal distribution
$\P_X$ on $(\cX,\sA)$ and the conditional distribution  $\P(\cdot|x)$ of a random variable $Y$ given 
$X=x$, see e.g., \citet[Thm. 10.2.1, p.\,343f]{Dudley2002}.

%

 
\section{Result}\label{result}

Let $(\O,\sA,\P)$ be a probability space and $(\cW,\tau_\cW)$ be a separable metric space equipped with the Borel $\s$-algebra $\B_\cW$. 
Denote the set of all $(\sA,\B_\cW)$-measurable functions  by $\mathcal{L}_0(\O,\cW)$ and the corresponding factor space of equivalence classes of functions, which are $\P$-almost everywhere identical, by $L_0(\O,\cW)$.
Then the Ky Fan metric defined on 
$\mathcal{L}_0(\O,\cW) \times \mathcal{L}_0(\O,\cW)$ is given by 
$$
   d_{KyFan}(f_1,f_2):= \inf \{\varepsilon \ge 0; \P(d_\cW(f_1,f_2)>\varepsilon)\le \varepsilon\} 
$$
for any $f_1,f_2\in \mathcal{L}_0(\O,\cW)$.
This metric metrizes convergence in probability, i.e., if 
$f,f_n \in L_0(\O,\cW)$, $n\in\N$, then $f_n \to f$ in probability if and only if
\begin{equation}
  \lim_{n\to\infty} d_{KyFan}(f_n,f) = 0,
\end{equation}
see e.g., \citet[Thm. 9.2.2]{Dudley2002}. 
Furthermore, $L_0(\O,\cW)$ is even complete for the Ky Fan metric,
if $(\O,\sA,\P)$ is a probability space and if $(\cW,d_\cW)$ a complete separable metric space, see \citet[Thm. 9.2.3, p. 290]{Dudley2002}. 
For our purpose it is more comfortable to consider equivalent metrics
and we will prove the next simple result to improve the readability of this note.

\begin{lemma}\label{lem.dpsimetric}
Let $(\O,\sA,\P)$ be a probability space, $(\cW,d_\cW)$ be a separable metric space, 
and $f, f_n: (\O,\sA)\to (\cW,\B_\cW)$ be measurable functions, $n\in\N$. 
Let $\psi:[0,\infty) \to [0,1]$
be a continuous, subadditive, and monotone increasing function with $\psi(0)=0$
and $\psi(x)>0$, if $x>0$, 
i.e., $\psi(x_1+x_2)\le \psi(x_1)+\psi(x_2)$,  and $\psi(x_1) \le \psi(x_2)$ for all $x_1,x_2\in[0,\infty)$ with $x_1 \le x_2$.
Then:
\begin{enumerate}
\item The function $d_\psi: L_0(\O,\cW)\times L_0(\O,\cW)\to [0,\infty)$ defined by
\begin{equation} \label{def.dpsi}
  d_\psi(f_1,f_2):=\int \psi\bigl(d_\cW(f_1,f_2)\bigr) \,d\P \, ,
  \qquad f_1, f_2 \in L_0(\O,\cW), 
\end{equation}
is a metric on $L_0(\O,\cW)$.
\item We have $f_n \to f$ in probability if and only if
\begin{equation} \label{conf.dpsi}
  \lim_{n\to\infty} d_\psi(f_n,f) = 0.
\end{equation}
\end{enumerate}
\end{lemma}

\begin{proofof}{\myem{Proof of Lemma \ref{lem.dpsimetric}}}
Part (i). Obviously, for any $f_1,f_2 \in L_0(\O,\cW)$ we have 
$d_\psi(f_1,f_2)=d_\psi(f_2,f_1) \ge 0$ and $d_\psi(f_1,f_2)=0$ if and only if
$f_1=f_2$, because $d_\cW$ is a metric and $\psi(x)>0$ if $x>0$. The triangle inequality for $d_\psi$ follows from the
triangle inequality for $d_\cW$ and the subadditivity of $\psi$. 
Hence $d_\psi$ is a metric.\\
Part (ii). Let us assume that $f_n \to f$ in probability. Then we have, for all
$\e>0$, that $\P(d_\cW(f_n,f)>\e)\to 0$, if $n\to\infty$. Because 
$\psi$ maps into the interval $[0,1]$ and $\psi$ is monotone increasing, it follows
\begin{eqnarray*} 
 & &  \psi\bigl(d_\cW(f_n,f)\bigr) \\
 & = & 
  \psi\bigl(d_\cW(f_n,f)\bigr) \cdot \eins_{\{d_\cW(f_n,f)>\e\}} + \psi\bigl(d_\cW(f_n,f)\bigr) \cdot \eins_{\{d_\cW(f_n,f)\le\e\}} \\
  & \le &
 1 \cdot \eins_{\{d_\cW(f_n,f)>\e\}} + \psi(\e) \cdot \eins_{\{d_\cW(f_n,f)\le\e\}} \, .
\end{eqnarray*}
Therefore,
\begin{eqnarray*}
 & & \int \psi\bigl(d_\cW(f_n,f)\bigr) \,d\P \\
 & \le &   
  \int \bigl( \eins_{\{d_\cW(f_n,f)>\e\}} + \psi(\e) \cdot \eins_{\{d_\cW(f_n,f)\le\e\}}\bigr) \,d\P  \\
 &  = &
  \P(d_\cW(f_n,f)>\e)+\psi(\e) \cdot \P(d_\cW(f_n,f)\le\e).
\end{eqnarray*}
Taking limits yields
$$
  0 \le \lim_{n\to\infty} \int \psi\bigl(d_\cW(f_n,f)\bigr) \,d\P 
  \le
  \psi(\e), \qquad \forall\,\e>0,
$$
which proves one direction, because $\psi$ is continuous and $\psi(0)=0$.\\
Let us now assume that  
$\int \psi\bigl(d_\cW(f_n,f)\bigr) \,d\P\to 0$, if $n\to\infty$.
The function $\psi$ is non-negative and monotone increasing by assumption.  
Hence 
\begin{eqnarray*}
  0 \le \psi(\e) \cdot \eins_{\{d_\cW(f_n,f)>\e\}} 
  \le    \psi\bigl(d_\cW(f_n,f)\bigr) \cdot \eins_{\{d_\cW(f_n,f)>\e\}} 
  \le    \psi\bigl(d_\cW(f_n,f)\bigr).
\end{eqnarray*}
Integrating with respect to $\P$ and then taking limits yields
$$
  0 \le \lim_{n\to\infty} \int \psi(\e) \cdot \eins_{\{d_\cW(f_n,f)>\e\}} \,d\P
  \le
   \lim_{n\to\infty} \int \psi\bigl(d_\cW(f_n,f)\bigr) \,d\P = 0,
   \quad \forall\,\e>0.
$$
Since $\psi(\e)>0$ for all $\e>0$, we conclude 
$\int \eins_{\{d_\cW(f_n,f)>\e\}} \,d\P = \P(d_\cW(f_n,f)>\e)\to 0$, if $n\to\infty$.
\qedr
\end{proofof}

Special cases are  
$\psi_1(x)=x/(1+x)$, see \citet[Thm. 17.1]{JacodProtter2004},
and $\psi_2(x)=\min\{1,x\}$, $x\ge 0$,
see e.g. 
\citet[Problem 9.2, p.\,353]{SC2008}, respectively.  The metric $d_{\psi_2}$ was used
to derive consistency in probability of support vector machines for kernel 
based quantile regression, see
\citet[Thm. 9.7, p. 343]{SC2008}.

We can now give our result which can be interesting for statistical
machine learning if the input set is $\cX$, the output space is $\cY$ 
and a function class $\mathcal{F}$ containing functions $f: \cX\to\cH$ are considered, where $\cH\subset \cY$. Special cases are $\cY=\cH$ and
$\cY=\cH=\R$.

\begin{theorem}\label{thm.convergence}
Let $(\cX,d_\cX)$ and $(\cY,d_\cY)$ be complete separable metric spaces 
and $(\cH,\|\cdot \|_\cH)$ be a separable Hilbert space with metric $d_\cH:=\|\cdot - \cdot\|_\cH$.  Equip these spaces with their
Borel $\s$-algebras $\B_\cX$, $\B_\cY$, and $\B_\cH$, respectively. Let 
$\P$ be a probability measure on $(\cX\times \cY, \B_{\cX\times \cY})$. 
Denote the set of all continuous functions $f:(\cX,d_\cX) \to (\cH,d_\cH)$
by $C(\cX,\cH)$.
Let $\mathcal{F}$ be a subset of $\mathcal{L}_0(\cX,\cH)$, where 
$\mathcal{F}$ is either a dense subset of $C(\cX,\cH)$ or $\mathcal{F}$ contains a dense subset of $C(\cX,\cH)$, where denseness is with respect to the metric $d_\psi$
defined in {(\ref{def.dpsi})}.\newline
Then $\mathcal{F}$ is dense in $\mathcal{L}_0(\cX,\cH)$ with respect to the metric $d_\psi$, i.e.,
for all $\varepsilon>0$ and for all $f\in \mathcal{L}_0(\cX,\cH)$ there exists 
$g_{\varepsilon,f} \in \mathcal{F}$ such that 
\be \label{dpsiineq}
d_\psi(f,g_{\varepsilon,f}) < \varepsilon. 
\ee
\end{theorem}

Please note, that the denseness notions in {(\ref{dpsiineq})} and in {(\ref{universalkernelineq})} differ: 
$d_\psi$ used in {(\ref{dpsiineq})} metrizes the convergence in probability for $\cH$-valued random quantities $f_n$, see Lemma \ref{lem.dpsimetric}, whereas
the much stronger supremum norm is used in {(\ref{universalkernelineq})}.

\begin{proofof}{\myem{Proof of Theorem \ref{thm.convergence}}}
Because $(\cY,d_\cY)$ is a complete separable metric space and hence a Polish
space, we can split the probability measure
$\P$ into its marginal distribution $\P_X$ and its conditional distribution
$\P(\cdot|x)$, $x\in \cX$.
Fix $\varepsilon>0$ and $f\in \mathcal{L}_0(\cX,\cH)$.
Lusin's theorem, see Theorem \ref{thm.LusinI}, gives the existence of a
closed set $\cX_{\varepsilon,f} \in \B(\cX)$ such that 
$$ 
  \P\bigl( (\cX\setminus \cX_{\varepsilon,f})\times \cY \bigr) = \P_X(\cX\setminus \cX_{\varepsilon,f}) < \frac{\varepsilon}{2}
$$
and the existence of a continuous function 
\begin{equation}
h_{\e,f}: (\cX_{\varepsilon,f}, {d_{\cX}}_{|\cX_{\varepsilon,f}}) \to (\cH,d_\cH) \nonumber
\end{equation}
such that 
\be \label{hequalsf}
  h_{\e,f}(x) = f(x), \qquad  x\in \cX_{\varepsilon,f}\,.
\ee
Because $h_{\e,f}$ is continuous, it is of course
$(\B_{\cX} \cap  \cX_{\varepsilon,f},\B_\cH)$-measurable. 
Recall that every normed space and in particular every Hilbert space is a Hausdorff locally convex space. 
Hence we can apply Dugundji's extension theorem, see Theorem \ref{thm.Dugundji}, which guarantees the existence of a
\emph{continuous} -- and therefore $(\B_\cX,\B_\cH)$-measurable -- function 
$$
  F_{\varepsilon,f}: (\cX,d_\cX) \to (\cH,d_\cH),
$$
such that 
\begin{equation}\label{fequalsF}
  F_{\e,f}(x)=h_{\e,f}(x) \qquad \forall \, x\in \cX_{\e,f}.
\end{equation}
Obviously, the continuous function $F_{\e,f}$ will in general not be identical to
the measurable function $f$.
Denote the indicator function of some set $A$ by $\eins_A$.
Because $\psi$ maps into the interval $[0,1]$, it follows
\begin{eqnarray*}
  d_\psi(f, F_{\varepsilon,f}) & = & 
  \int \psi\bigl(d_\cH(f,F_{\varepsilon,f})\bigr) \,d\P \\
  & = & \int \psi\bigl(d_\cH(f,F_{\varepsilon,f})\bigr) \eins_{\cX_{\varepsilon,f}}\,d\P 
        + \int \psi\bigl(d_\cH(f,F_{\varepsilon,f})\bigr) \eins_{\cX\setminus \cX_{\varepsilon,f}}\,d\P \\
  & \stackrel{\scriptsize{(\ref{fequalsF}), (\ref{hequalsf})}}{=} & \int \psi\bigl(d_\cH(f,f)\bigr) \eins_{\cX_{\varepsilon,f}}\,d\P 
        + \int \psi\bigl(d_\cH(f,F_{\varepsilon,f})\bigr) \eins_{\cX\setminus \cX_{\varepsilon,f}}\,d\P \\
  & \stackrel{\scriptsize{\psi(0)=0,~ \inorm{\psi}\le 1}}{\le} & \int 0 \cdot  \eins_{\cX_{\varepsilon,f}}\,d\P  + \int 1 \cdot \eins_{\cX\setminus \cX_{\varepsilon,f}}\,d\P \\
  & = & 
   \P(\cX\setminus \cX_{\varepsilon,f}) ~ < ~ \frac{\varepsilon}{2}\, .
\end{eqnarray*}

If the continuous function $F_{\varepsilon,f}$ is an element of $\mathcal{F}$, then we can choose
$g_{\varepsilon,f}=F_{\varepsilon,f}$ and we have
$d_\psi(f,g_{\varepsilon,f}) < \frac{\e}{2}$.
If the continuous function $F_{\varepsilon,f}$ is not an element of $\mathcal{F}$, then the denseness assumption of  
$\mathcal{F}$ guarantees the existence of a continuous function
$g_{\varepsilon,f} \in \mathcal{F}$ such that 
$d_\psi(F_{\varepsilon,f}, g_{\varepsilon,f}) < \frac{\varepsilon}{2}$.
We then obtain
$d_\psi(f,g_{\varepsilon,f}) < \e$ by 
the triangle inequality,
which completes the proof.
\qedr
\end{proofof}

\begin{example}
Let $k:\cX\times \cX\to \R$ be a universal kernel with RKHS $H$, where
$(\cX,d_\cX)$ is a compact metric space.
Then, for all $f\in C(\cX,\R)$ and for all $\varepsilon>0$, 
there exists $g_{\varepsilon,f} \in H$ such that 
$\inorm{f-g_{\varepsilon,f}} < \varepsilon$.
As convergence with respect to the supremum norm implies 
convergence in probability, we immediately obtain that 
for all $f\in C(\cX,\R)$ and for all $\varepsilon>0$ 
there exists $\tilde{g}_{\varepsilon,f} \in H$ such that 
$d_\psi(f,\tilde{g}_{\varepsilon,f}) < \varepsilon$.
One special case is the Gaussian RBF-kernel $k_\gamma$
with bandwidth $\gamma>0$ defined on some compact set $\cX\subset \R^d$. 
This kernel is well-known to be universal,
see e.g., \citet[Cor. 4.58]{SC2008}. 
Another special case is the universal kernel 
$k_\s$ defined on the set of all Borel probability measures, see
\citet[Example 1]{ChristmannSteinwart2010nips} for details.
Let $\cX=\mathcal{M}_1(\Omega,\B(\O))$, where $(\Omega,d_\Omega)$ is some compact metric space and $k_\Omega$ is a continuous kernel on $\Omega$ with canonical feature map 
$\Phi_\Omega$ and RKHS
$H_\Omega$. Assume that $k_\Omega$ is a so-called characteristic kernel in the sense that the function $\rho: X \to H_\Omega$ defined by 
$\rho(P)=\Ex_\P \Phi_\Omega$ is injective.
Then the Gaussian-type RBF-kernel 
$$
k_\s(\P,\P'):=\exp\bigl( -\frac{1}{\gamma^2} \| \Ex_\P \Phi_\Omega -
\Ex_{\P'} \Phi_\Omega  \|_{H_\Omega}^2 \bigr), \qquad 
\P,\P'\in \mathcal{M}_1(\Omega,\B(\O)),
$$
is a universal kernel on $\cX$ and obviously even bounded. 
\end{example}
 
%
 
        
%

 
\begin{example}
Let $\cX$ be a complete separable metric space and $\cY = [-M,+M]$ for some fixed 
constant $M\in(0,\infty)$. Let $L$ be a convex and Lipschitz continuous loss fucntion with Lipschitz constant $|L|_1>0$. Consider the minimizer $f_{L,\textup{D},\lambda_n}$ defined by minimizing (\ref{SVMs}).

If $f_0 \in \arg\min\left\{\mathcal R_{L,\P}(f) \mid f\in\mathcal L_0(\mathcal X, \mathcal Y)\right\}$ exists, it follows directly that $f_0(x) \in [-M,+M]$ for all $x\in\cX$. Therefore it is natural to project $f_{L,\textup{D},\lambda_n}$ onto $[-M,+M]$ obtaining
$$\hat{f}_{L,\textup{D},\lambda_n} := \textup{max}\left\{-M, \textup{min}\left\{+M, f_{L,\textup{D},\lambda_n}\right\}\right\},$$
see e.g., \citet[Section 10.2]{CuckerZhou2007}. Hence, let us define  
$$
\mathcal F := \bigl\{g =  \textup{max}\left\{-M, \textup{min}\left\{+M, f\right\}\right\}\mid f\in H\bigr\},$$
where $H$ is the RKHS corresponding to the chosen kernel $k$.

If  $\mathcal F$ is dense in $\mathcal L_0(\cX,\cY)$ with respect to $d_\psi$, then, for $f_0\in\mathcal F$ and for all $\varepsilon>0$, there exists 
a sequence $(g_n)_{n\in\mathds N}\subset\mathcal F$ and a positive integer $n_0$ such that for all $n\ge n_0$ it holds $d_\psi(g_n,f_0) < \varepsilon$. Hence Lemma \ref{lem.dpsimetric} implies that $g_n \to f_0$ in probability for $n\to\infty$.

Under the conditions that $(g_n)_{n\in\mathds N}\subset\mathcal F$ and therefore $|g_n(x)| \le M$ for all $n\in\mathds N$ and all $x\in\cX$, it holds true that 
\begin{align}\label{L1norm}
  \lim_{n\to\infty} \left\|g_n-f_0\right\|_{L_1(\P_X)} = 0
\end{align}
for all marginal distributions $\P_X$ on $\cX$.
Recalling the Lipschitz continuity of the loss function $L$ we have, see e.g., \citet[Lem.~2.19]{SC2008},
\begin{align}\label{riskdiff}
\left|\mathcal R_{L,\P}(f) - \mathcal R_{L,\P}(f_0)\right|\le |L|_1 \left\|f-f_0\right\|_{L_1(\P_X)} \text{\ for all\ } f\in\mathcal F.
\end{align}
Combining (\ref{L1norm}) and (\ref{riskdiff}) we obtain
$$\underset{n\to\infty}{\textup{lim}} \mathcal R_{L,\P}(g_n) = \mathcal R_{L,\P}(f_0),$$
where $g_n \in\mathcal F$ for all $n\in\mathds N$. 
\end{example}


\end{document}